\documentclass{esannV2}
\usepackage{graphicx}
\usepackage[latin1]{inputenc}
\usepackage{amssymb,amsmath,array}
\usepackage{makecell}

\begin{document}
\title{Sawtooth Sampling for Time Series Denoising Diffusion Implicit Models}

\author{Heiko Oppel$^1$, Andreas Spilz$^1$ and Michael Munz$^1$
%
%
\vspace{.3cm}\\
%
1- AI for Sensor Data Analytics Research Group,
  Ulm University of Applied Sciences
%
}

\maketitle

\begin{abstract}
Denoising Diffusion Probabilistic Models (DDPMs) can generate synthetic timeseries data to help improve the performance of a classifier, but their sampling process is computationally expensive. 
We address this by combining implicit diffusion models with a novel Sawtooth Sampler that accelerates the reverse process and can be applied to any pretrained diffusion model. 
Our approach achieves a 30 times speed-up over the standard baseline while also enhancing the quality of the generated sequences for classification tasks.
\end{abstract}

\section{Introduction}
Ho et al. \cite{ddpm_ho_paper_2020} simplified the training objective for Diffusion Models and made them feasible for generating images from pure gaussian noise. 
Subsequent developments improved Denoising Diffusion Probabilistic Models (DDPMs) with the deterministic Denoising Diffusion Implicit Model (DDIM) formulation \cite{song2021ddim} and demonstrated that noise scheduling and non-Markovian sampling can dramatically accelerate inference, often without sacrificing fidelity.
Other strategies e.g. from Jiang et al. \cite{jiang2025fastddpmfastdenoisingdiffusion} reduced both training and sampling time by learning on a restricted set of time steps, while Xu et al. \cite{xu2023restartsamplingimprovinggenerative} introduced Restart Sampling, which interleaves forward and backward steps to reduce function evaluations and improved Frechet Inception Distance.
\\
While most advances target image synthesis, diffusion models have been successfully adapted to audio and broader time series domains. 
Several groups even explored Inertial Measurement Unit (IMU) data generation like \cite{10157482} and \cite{OppelDDPM}, though without focusing on inference speed.
\\
Our work builds directly on \cite{OppelDDPM} for IMU data  and their C-Opt GAK similarity metric \cite{oppel2025timeseriessimilarityscore}, which we use to track generative quality throughout the denoising process.
To accelerate sampling, we leverage the deterministic DDIM framework \cite{song2021ddim} and introduce a novel Sawtooth Sampler. 
Inspired by the Restart approach \cite{xu2023restartsamplingimprovinggenerative} yet omitting additional forward steps and extra noise, our sampler reduces the required denoising steps by a factor of 30 while further suppressing high-frequency artifacts. 
Coupled with the similarity score, we were able to  analyze the denoising process at each time step.

\section{Methodology}
\subsection{Datasets}
We used two timeseries datasets: one with four cyclic human movements and one with climbing activities for fall detection via an instrumented belay device. 
All data were recorded with a single IMU, supplemented by Hall-Sensors for rope tracking in the climbing dataset. 
The IMU was placed on the right thigh for HAR and on the belay device for climbing. 
The climbing data is showing a 1:9 fall-ascent ratio as an ascent takes longer than a fall. 
The HAR dataset contains 12 participants, the climbing dataset includes 37 falls and 19 ascents.
They are further described in \cite{oppel2025timeseriessimilarityscore}.

\subsection{Denoising Diffusion Implicit Model}
The DDIM replaces the DDPMs inference step by using a non-Markovian diffusion process.
The formula to generate a sample using the DDIM \cite{song2021ddim} approach changes to:
\begin{align}\label{eq:ddim_sampling}
    x_{\tau_{i-1}}  \sqrt{\bar{\alpha}_{\tau_{i-1}}} \left(\frac{x_{\tau_i} - \sqrt{1-\bar{\alpha}_{\tau_i}} \epsilon_\theta^{(\tau_i)}(x_{\tau_i})}{\sqrt{\bar{\alpha}_{\tau_i}}}\right) + \sqrt{1-\bar{\alpha}_{\tau_{i-1}} - \sigma_{\tau_i}^2} \epsilon_{\theta}^{(\tau_i)}(x_{\tau_i}) + \sigma_{\tau_i}\epsilon_{\tau_i} 
\end{align}
with $\tau_i$ being a time step in the denoising process $\forall i \in S$ and $S$ being a subsequence of $t=1, ...,T$ and $\tau_S=T$.
We have chosen different start and end values for the diffusion rate $\beta_t=1-\alpha_t$ depending on the sensor type. 
Nevertheless, they follow a linear function.
Note, that it holds $\bar{\alpha}_t = \prod_{s=1}^{t}\alpha_s$ and $\epsilon_{\theta}^{\tau_i}(x_{\tau_i})$ being the noise at time $\tau_i$ given $x_{\tau_i}$ as well as $\epsilon_{\tau_i} \sim \mathcal{N}(0, \mathbb{I})$ which is independent of $x(\tau_i)$. 
One major difference between DDPM and DDIM is the stochasticity of the sampling process.
The DDIM is deterministic in a way, that $x_0$ is only dependent on the initial state $x_T$, if the control parameter for the stochasticity $\sigma_{\tau_i} = 0, \forall i \in S$ \cite{song2021ddim}, which we chose for the denoising process.

\subsubsection{Sawtooth Sampling}
Figure 1 illustrates the Sawtooth Sampling method. We begin with the standard DDIM reverse process (Eq. 2), then reset the variance scheduler and restart denoising after each of $N$ resampling steps. 
We test $N \in \{1, 2, 5, 10\}$, referring to these variants as DDIM-KN, with $N=1$ corresponding to standard DDIM. 
To ensure fair comparison, we fixed the total number of denoising steps to 100 for all variants and datasets, distributing them evenly across the $N$ iterations.
The complete process is defined as:
\begin{equation}
    x_{\tau_{i-1}}^{k} = \text{DDIM}(x_{\tau_{i}}^{k}), \text{with } k=1, 2, ..., N
\end{equation}

\begin{figure}[h]
    \centering
    \includegraphics[width=0.6\linewidth]{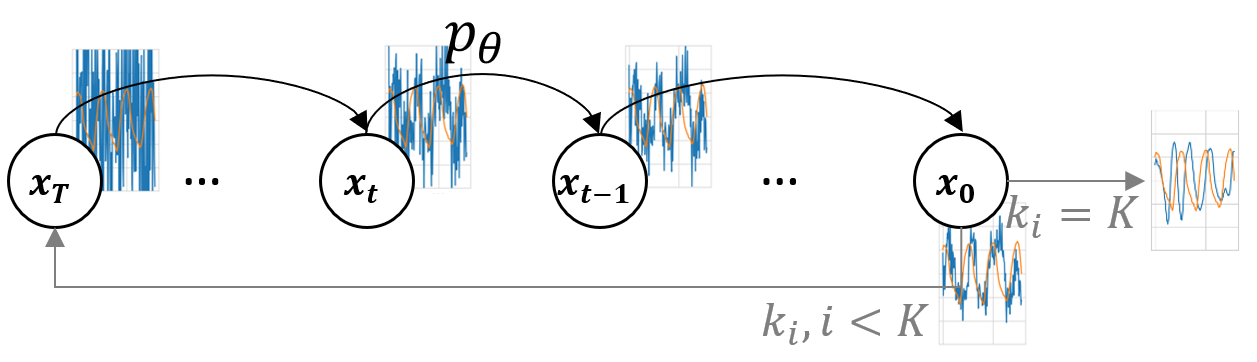}
    \caption{Illustration of the sampling process using the Sawtooth sampler.}
    \label{fig:sawtooth_sampler_illustration}
\end{figure}

\subsubsection{Variance Scheduler}
We used a linear function for the scheduler function, though any function can be used.
As the variance $\beta_t$ is limited to a pre-defined range which is initially set, it is guaranteed that the criteria $0 < \beta_t < 1$ is still met, so that the mean of the conditional distribution $q(x_t|x_0)$ converges to $0$ for $t \rightarrow \infty$ in the forward process for all $k$ iterations.

\subsection{Evaluation Methods}
We evaluated the Sawtooth Sampler using two metrics: a CNN classifier and a similarity measure comparing each generated sequence to its most similar real counterpart. 
The classifier uses two convolution-max-pool blocks followed by a dense network. 
For HAR, models were trained under LOSO using a reduced dataset (two samples per participant and class), except for the Full-Set model. 
For DDPM and DDIM, ~15,000 synthetic sequences were added to training. 
For the Climbing dataset, we used two baselines: one trained on downsampled real data to compensate for the class disbalance and one trained solely on DDPM-generated data.
The similarity metric, designed for time series data, requires pretraining on the training and validation sets to establish a ground truth about the within-class similarity. 
It can then be used to identify the closest real sequence for each synthetic sample including a score value. 
Scores range from 0 (no similarity) to 1 (identical).
The classifier architecture as well as the similarity metric and the baseline models are in more detail explained in \cite{OppelDDPM} and \cite{oppel2025timeseriessimilarityscore}.

\section{Results}
\subsection{Effect of the Sawtooth Sampler on the Denoising Process}
To analyse the generated sequences, we applied the C-Opt GAK similarity score of Oppel et al. \cite{oppel2025timeseriessimilarityscore}, which quantifies agreement between the power spectral densities of real and generated sequences. 
We evaluated each generated sequence at each denoising step for each Sawtooth configuration (DDIM-K1, K2, K5, K10) and compared it to its most similar real sequence. 
Representative results for one class from each dataset are shown in Figure 2.

Across both datasets, the denoising curves exhibit multiple local maxima aligned with the k resampling steps. 
For HAR data, these maxima are subtle for K5 and K10 in early steps but become more pronounced with later steps. 
In most cases, the trend of the similarity score either increased or remained stable across resampling steps.
However, in HAR K10 occasionally showed an early increase followed by a decline, returning lesser similarity levels than with any other Sawtooth configuration by the final step. 

\begin{figure}[h]
    \centering
    \includegraphics[width=0.75\linewidth]{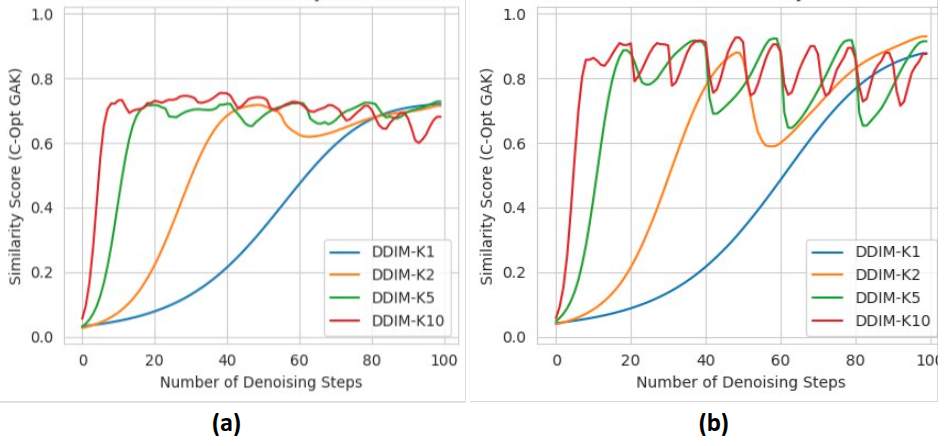}
    \caption{C-Opt GAK similarity scores across denoising steps for each DDIM Sawtooth configuration: (a) Cycling class in the HAR dataset (averaged over PID3 sequences) and (b) AC21 Fall class in the Climbing dataset with its closest synthetic match.}
    \label{fig:sim_scores}
\end{figure}

\subsection{Time Improvement with DDIM}
We reduced the denoising steps from 3000 to 100, hence directly lowering the generation time. 
On an NVIDIA GeForce RTX 3090, producing 128 samples took about 5 minutes 30 seconds with DDPM, but only $\sim$11 seconds using our DDIM-based approach.

\subsection{Classification Results}
\begin{figure}[h]
    \centering
    \includegraphics[width=0.6\linewidth]{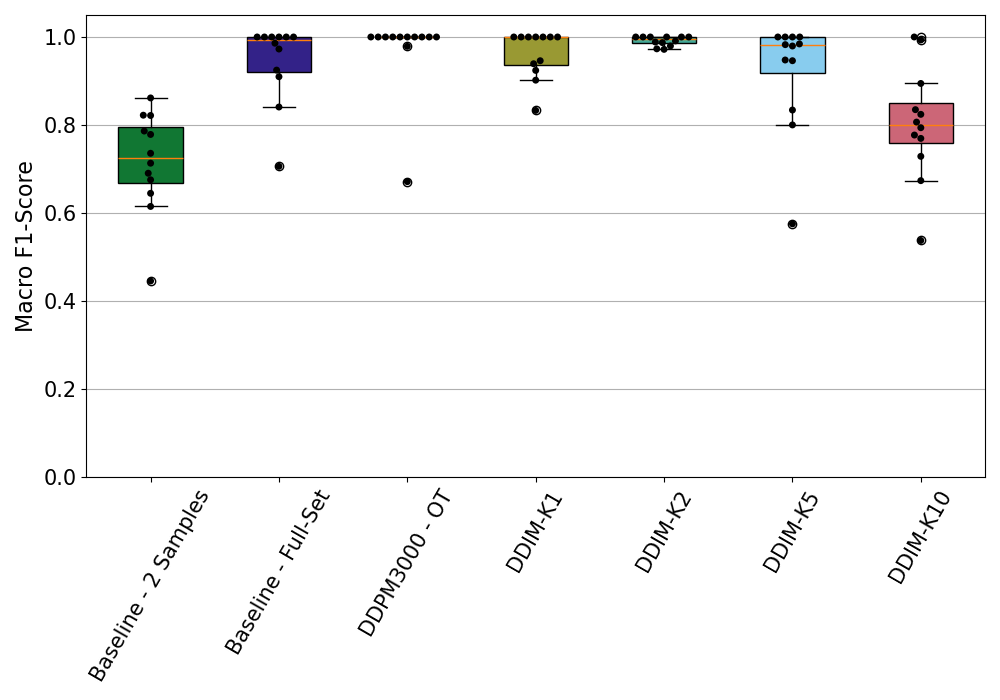}
    \caption{
    Macro F1-score for each LOSOCV run on the HAR dataset, comparing all denoising methods to the baseline. Each point represents a participant used as the held-out test set.
    }
    \label{fig:classif_har}
\end{figure}

With one out of the four Sampler configurations we achieved a comparable result to the best baseline model for the HAR dataset, see Figure~\ref{fig:classif_har}.
DDIM-K1 achieved a macro F1-score of 1.0 for 7 of 12 participants, but dropped below 0.95 for the remaining 5. 
DDIM-K2 delivered the strongest overall performance, with the lowest score occurring on PID3 (0.97). 
Increasing K beyond 2 generally reduced the classifier performance, most notably for K=10, where only PID5 reached an F1-score of 1.0. 
DDIM-K2 outperformed all real-data baselines, though the DDPM3000 Optimal Control model slightly exceeded it for 5 participants (by up to 0.028) and performed worse for 2 participants (0.30 and 0.021).\\
With the climbing dataset, we not only matched the baseline performance but exceeded it, see Table~\ref{tab:climbing_results}.
Using only real data, downsampling the minority class lead to the best baseline results on real data only, with geometric means of $0.858\pm0.029$ (3-class) and $0.927\pm0.027$ (2-class). 
Adding DDPM generated data improved these averages by $0.034$ and $0.004$, respectively. 
Standard DDIM sampling (DDIM-K1) reduced the performance relative to this baseline, but all Sawtooth Sampling variants improved it, with the 2 step Sawtooth Sampler achieving the strongest gains.

\begin{table}[h]\tiny
    \centering
    \caption{
    Classification results using the climbing dataset.}
    \resizebox{\textwidth}{!}{\begin{tabular}{lcccccc}
        \hline    
          \makecell{Evaluation \\Metric} & \makecell{TRTR \& \\ Downsampling} & \makecell{TSTR \\ DDPM3000} &  \makecell{TSTR \\ DDIM-K1} &  \makecell{TSTR \\ DDIM-K2} &  \makecell{TSTR \\ DDIM-K5} &  \makecell{TSTR \\ DDIM-K10} \\
        \hline
        \makecell{3 Class \\Geometric Mean} & $0.858 \pm 0.029$ & $0.892 \pm 0.030$ & $0.770 \pm 0.026$ & $0.913 \pm 0.037$ & $0.884 \pm 0.025$ & $0.871 \pm 0.039$\\
        
        \makecell{Fall vs Climbing\\Geometric Mean} & $0.927 \pm 0,027$ & $0.931 \pm 0.031$ & $0.925 \pm 0.023$ & $0.949 \pm 0.033$ & $0.938 \pm 0.023$ & $0.930 \pm 0.030$ \\
        \hline
    \end{tabular}}
    \label{tab:climbing_results}
\end{table}

\section{Discussion}
Although DDIM is known to accelerate DDPM sampling, its performance on timeseries data, especially IMU signals dominated by high frequency noise can be limited. 
In our experiments, standard DDIM removed noise less effectively than the full DDPM reverse process, which relies on thousands of steps. 
In contrast, our Sawtooth Sampler improves high frequency noise suppression without increasing computational cost.
A comparison with Xu et al.'s Restart approach \cite{xu2023restartsamplingimprovinggenerative} would be valuable. 
Their method injects additional noise before restarting the reverse process, which may be harder to remove in IMU data and could behave similarly to our DDIM-K1 results despite its success in image domains. 
Restart sampling also introduces extra hyperparameters $[t_{min}, t_{max}]$, whereas our method avoids such tuning.\\
The Sawtooth Sampler periodically resets the denoising trajectory, producing intermediate states that deviate from the standard normal distribution at each resampling step $k>1$. 
Despite this, the deterministic nature of DDIM and the noisy characteristics of the underlying data allow the model to continue removing high-frequency noise. 
Stochasticity enters only through the initialization of $x_0$.
A clear limitation emerges with excessive resampling. 
For $K=10$, the model often reintroduced high-frequency noise after previously suppressing it. 
As suggested by Equation~\ref{eq:ddim_sampling}, once the synthetic sequences noise level approaches that of the real data, the model lacks a meaningful noise reference and may estimate residuals that inadvertently amplify high frequency components.
When the number of resampling steps is kept moderate, however, the Sawtooth Sampler consistently removes high frequency noise more effectively than the standard sampler. 

\section{Conclusion}
In this paper, we introduced the Sawtooth Sampler to enhance DDIM-generated sequences, using two movement datasets: HAR and Climbing.
We evaluated multiple Sampling configurations and identified their limitations. 
Excessive resampling reintroduced high frequency noise, leading to less realistic outputs. 
The C-Opt GAK similarity metric allowed us to detect this effect both visually and objectively, and these findings aligned with the classification results obtained from models trained on each Sawtooth Sampling variant.

\section*{Data Availability Statement}
The HAR data that support the findings of this study is available at the UCI Machine Learning Repository\\(https://archive.ics.uci.edu/dataset/305/realdisp+activity+recognition+dataset) and was originally collected by Oresti Banos et al.:
Banos, Oresti, Mate Toth, and Oliver Amft. "REALDISP Activity Recognition Dataset." UCI Machine Learning Repository, 2012, https://doi.org/10.24432/C5GP6D.

\section*{Funding}
This work has been partially funded by the Carl-Zeiss-Stiftung.

\bibliographystyle{unsrt}
\bibliography{references.bib}


\end{document}